\documentclass[11pt,a4paper]{article}
\usepackage[hyperref]{acl2019}
\usepackage{times}
\usepackage{latexsym}
\usepackage{url}
\usepackage[colorinlistoftodos]{todonotes}
\usepackage{array}
\usepackage{xspace}
\usepackage{microtype}
\usepackage{graphicx}
\usepackage[labelformat=simple]{subcaption}

\usepackage{amsmath}

\usepackage{titlesec}

\usepackage{microtype}

\newcommand{\dataset}[1]{#1\xspace}
\newcommand{\wattack}{\dataset{W-ATT}}
\newcommand{\wtoxic}{\dataset{W-TOX}}
\newcommand{\hatespeech}{\dataset{Hatespeech}}
\newcommand{\fastText}{\dataset{fastText}}

\aclfinalcopy 
\def\aclpaperid{32} 

\title{Neural Word Decomposition Models for Abusive Language Detection}

\author{Sravan Babu Bodapati  \quad Spandana Gella  \\
\quad \bf{Kasturi Bhattacharjee} \quad \bf{Yaser Al-Onaizan} \\
\quad \quad \textbf{Amazon AI, USA} \\ 
{\{sravanb, sgella, kastb, onaizan\}@amazon.com}}

\date{}

\begin{document}
\maketitle
\begin{abstract}

User generated text on social media often suffers from a lot of
undesired characteristics including hatespeech, abusive language, 
insults etc. that are targeted to attack or abuse a specific group 
of people. Often such text is written differently compared to traditional 
text such as news involving either explicit mention of abusive words, obfuscated words and typological errors or implicit abuse i.e., indicating or targeting negative stereotypes. Thus, processing this text poses several robustness challenges when 
we apply natural language processing techniques developed for traditional text.
For example, using word or token based models to process such text can treat two spelling variants of a word as two different words. Following recent work, 
we analyze how character, subword and byte pair encoding (BPE) models can be 
aid some of the challenges posed by user generated text. 
In our work, we analyze the effectiveness 
of each of the above techniques, compare and contrast various  
word decomposition techniques when used in combination with others. We experiment with finetuning large pretrained language models, and demonstrate their robustness 
to domain shift by studying Wikipedia attack, toxicity 
and Twitter hatespeech datasets.

\end{abstract}

\section{Introduction}
\label{sec:intro}

In recent years, with the growing popularity of social media applications, there has been an exponential increase in the amount of user-generated text 
including microblog posts, status updates and comments posted on the web. 
The power of communicating freely with large number of users has 
resulted in not only sharing news and exchanging content but 
has also led to a problem of large number of harmful, offensive and 
aggressive 
interactions online \cite{duggan2017online}. Previous work on 
identifying abusive language has tackled this problem by training 
computational methods that are capable of automatically recognizing offensive content for text on MySpace \cite{yin2009detection}, Twitter \cite{waseem2016hateful,davidson2017automated}, Wikipedia comments \cite{wulczyn2017ex} and Facebook posts \cite{VignaCDPT17,kumar2018benchmarking}.

Most of these models are based on features extracted from words or word n-grams or the recurrent neural networks that operate on word embeddings \cite{pavlopoulos2017deep,badjatiya2017deep} with few exceptions of models that utilize character n-grams that can model noisy text and out-of-vocabulary words \cite{waseem2016hateful,nobata2016abusive,wulczyn2017ex}. However, these models are not very effective at modeling 
obfuscated words such as \textit{w0m3n, nlgg3r} which are prominent 
in user generated text that are intended at evading hate speech detection \cite{mishra2018neural}. In this work, we aim to address this by investigating word, subword and character-based models for abusive language detection.

Recent advances in unsupervised pre-training of 
language models have led to strong improvements on various general natural 
language processing and understanding tasks such as question answering, sentiment and natural language inference \cite{elmo,devlin2018bert}. However, it is unclear how such models trained on standard text would transfer information when fine-tuned on noisy user generated text. In additional to studying word, subword and character-based model performances on abusive language detection we also combine these with pre-trained embeddings and fine-tuning these pre-trained language models and understand their efficiency and robustness in identifying abusive text.

Specifically, in this work, we address the 
effectiveness of character-based models, subword or Byte Pair Encoding (BPE) based models 
and word features based models along with pre-trained word embeddings 
and fine tuning pretrained language models for detecting abusive language in text. 
Precisely we make following contributions:

\begin{itemize}
    \item We compare the effectiveness of end-to-end character based models, with word + character embedding models, byte pair encoding and subword models, to show which of the techniques perform better than pure word based models.
    \item We demonstrate how fine-tuning large pre-trained language models, the latest breakthrough in NLP, enhance state of the art on few of the abusive language datasets, and show that the domain shift isn't considerable when applied to abusive language datasets.
    \item We also examine how preprocessing documents with byte pair encoding model pretrained on a large corpus, boost the performance of several word embedding based models massively.
\end{itemize}


\section{Related Work}

Identifying abusive context on the web is one of the widely studied topics on social media text.
This problem has been studied for Hate Speech detection \cite{kwok2013locate, waseem2016hateful, waseem2016you, ross2016hatespeech,saleem2017hate,warner2012detecting}, Harassment \cite{yin2009detection,cheng2015antisocial}, Cyberbullying \cite{willard2007cyberbullying,tokunaga2010following,schrock2011problematic}, Abusive language detection \cite{sahlgren2018learning,nobata2016abusive}, aggression identification \cite{kumar2018benchmarking,aroyehun2018aggression,modha2018filtering}, identifying toxic comments on forums \cite{wulczyn2017ex} and offensive language identification \cite{wiegand2018overview,zampieri2019semeval}. While most of the work in identifying abusive on social media is predominantly studied for English social media posts some of the latest work include study on German \cite{wiegand2018overview}, Italian \cite{bosco2018overview} and Mexican Spanish \cite{alvarez2018overview}.


Some of the early methods on identifying abusive text used word n-gram, part-of-speech (POS) tagging (syntactic features), manually created profanity lexicons or stereotypical words, TF-IDF features along with sentiment and contextual features and trained supervised classifiers such as support vector machines. \cite{yin2009detection,warner2012detecting}. \citet{waseem2016you} studied character n-grams, skipgrams, brown clusters and POS tag based features for identifying hatespeech. \citet{waseem2016hateful} studied usefulness of various socio-linguistic features such as gender, location, word-length distribution, Author Historical Salient Terms (AHST) features in identifying hatespeech.

Some of the recent work compared efficiency of both character n-gram based models as inputs to logistic regression and multi-layer perceptron models \cite{wulczyn2017ex}. \citet{nobata2016abusive} showed that character n-grams features alone can perform well and can efficiently model noisy text. They also showed off-the-shelf word embeddings can be used to identify abusive text. 

\citet{pavlopoulos2017deep} used deep-learning based models specifically they employed RNN with a novel classification-specific attention mechanism and achieve state-of-the-art results on identifying attack and toxic content in Wikipedia comments. \citet{badjatiya2017deep} investigated three different neural networks for hatespeech detection: (i) Convolutional neural network (inspired by CNN's for sentiment classification by \citet{kim2014convolutional}) (ii) Long short-term memory networks (LSTM) to capture long range dependencies and (iii) FastText classification model that represents document by averaging word vectors that can be fine-tuned for the hatespeech task.

While \citet{badjatiya2017deep} analyzed various architectures to encode text for hatespeech detection, we are not aware of any work that studied various word decomposition models for identifying abusive language in text. Recent work on identifying offensive language in text include fine-tuning large pretrained languege model BERT which use subword units to encode text \cite{zampieri2019semeval,zhu2019iu}.  For the SEMEVAL-2019 task of 
offensive language identification 7 out of top 10 submissions used BERT finet tuning.  \citet{zampieri2019semeval} highlighted that 8\% of 104 systems participated in the shared task used BERT based fine-tuning. 

In this work, we analyze the effectiveness of different ways of learning representations with character-based models, subword or BPE based models and word features based models. We also combine these with well known pre-trained word embeddings and very large pretrained language models for fine-tuning and detecting abusive language in text. In Section ~\ref{sec:datasets} we describe the datasets that we study in this work for hatespeech and abusive detection.

\section{Datasets}
\label{sec:datasets}

We experiment with three datasets: Twitter dataset \cite{waseem2016hateful}, 
Personal Attack and Toxicity datasets from Wikipedia Talk 
dataset \cite{wulczyn2017ex} that covers sexism/racism, 
personal attack and toxicity aspects of abuse in user generated text online. 

\subsection{Twitter Dataset}
\label{ssec:twitter-dataset}

We use the hatespeech Twitter dataset (\hatespeech) provided by \citet{waseem2016hateful}. 
This dataset was created from a corpus of 136k tweets collected 
from Twitter by searching for commonly used racist and sexist slurs 
on various ethnic, gender and religious minorities over a two-month period. 
The original data had 16,907 tweets corresponding to sexist, racist and 
neither labels (3378, 1970 and 11559 respectively). 
However, we could only retrieve 11170 of the tweets (2914: sexism, 
17: racism and 8239: neither) with python's Tweepy 
library. 
Similar issue of missing tweets has been reported by \citet{mishra2018neural}. However, the percent of tweets we lost are much higher than theirs and most of the tweets lost are for the \textit{racism} label. 
We have lost majority of the tweets 
corresponding to sexism label. Since we lost large chunk of tweets we conduct our experiments on cross validation of 5 splits and report scores on all of the 5 splits. 

\subsection{Wikipedia talk page}
\label{ssec:wiki-talk-datasets}

We use the personal attacks (\wattack) and toxicity (\wtoxic) datasets 
that were randomly sampled from 63 Million talk page comments from the 
public dump of English Wikipedia by \citet{wulczyn2017ex}. 
Each comment in both the datasets were annotated by at least 
10 workers and we use the majority label as its gold label. 
Overall, we have 115.8k comments in \wattack dataset 
(69.5k, 23.1k and 23.1k in train, dev and test splits respectively) 
and 159.6k comments in \wtoxic dataset (95.6k, 32.1k and 31.8k in train, 
dev and test splits). Similar to hatespeech dataset both 
the \wattack and \wtoxic datasets also have skewed distribution 
of labels having 13.5\% and 15.3\% of them labeled as abusive.

\section{Methods}
\label{sec:methods}

In this section, we present various word decomposition methods and modeling architectures we analysed for studying Twitter and Wiki Talk page \wattack and \wtoxic comment datasets.


\subsection{Word-based Model}
\label{ssec:word-based-model}

As a baseline, we adpot the fastText \cite{GraveMJB17} classification algorithm. The fastText algorithm performs mean pooling on top of the word embeddings $w_i^{emb}$ to obtain a document representation. This document representation is passed through a Softmax layer to obtain classification scores. The embeddings can either be learned or can be initialized with pre-trained embeddings and fine-tuned during training. We run multiple variants of fastText in our experiments. 

\subsection{Subword-based Model}
\label{ssec:subword-models}
 Subwords are formed by concatenating all the characters of a particular length within a word boundary. Addition of subwords gives the model ability to learn words which are misspelled such as \textit{emnlp} and \textit{emnnlp} are similar. A pure word based model would consider \textit{emnnlp} as out-of-vocabulary (OOV) word, if not seen in training set, but a subword model would decompose \textit{emnnlp} into ``emn'' and ``nlp'', and train subword embeddings $w_{sub}^{emb}$  for each of these subwords. We take subword variant of fastText model to incorporate subword context into the model. The algorithm considers all subwords of varying lengths within the boundary of a word. 

\subsection{Joint Word and Character Embedding Model}
\label{ssec:word-char-embeddings}



\begin{figure*}[!ht]
     \begin{minipage}[l]{1.0\columnwidth}
         \centering
         \includegraphics[trim=1.5cm 2cm 1.5cm 4cm,clip,width=\textwidth]{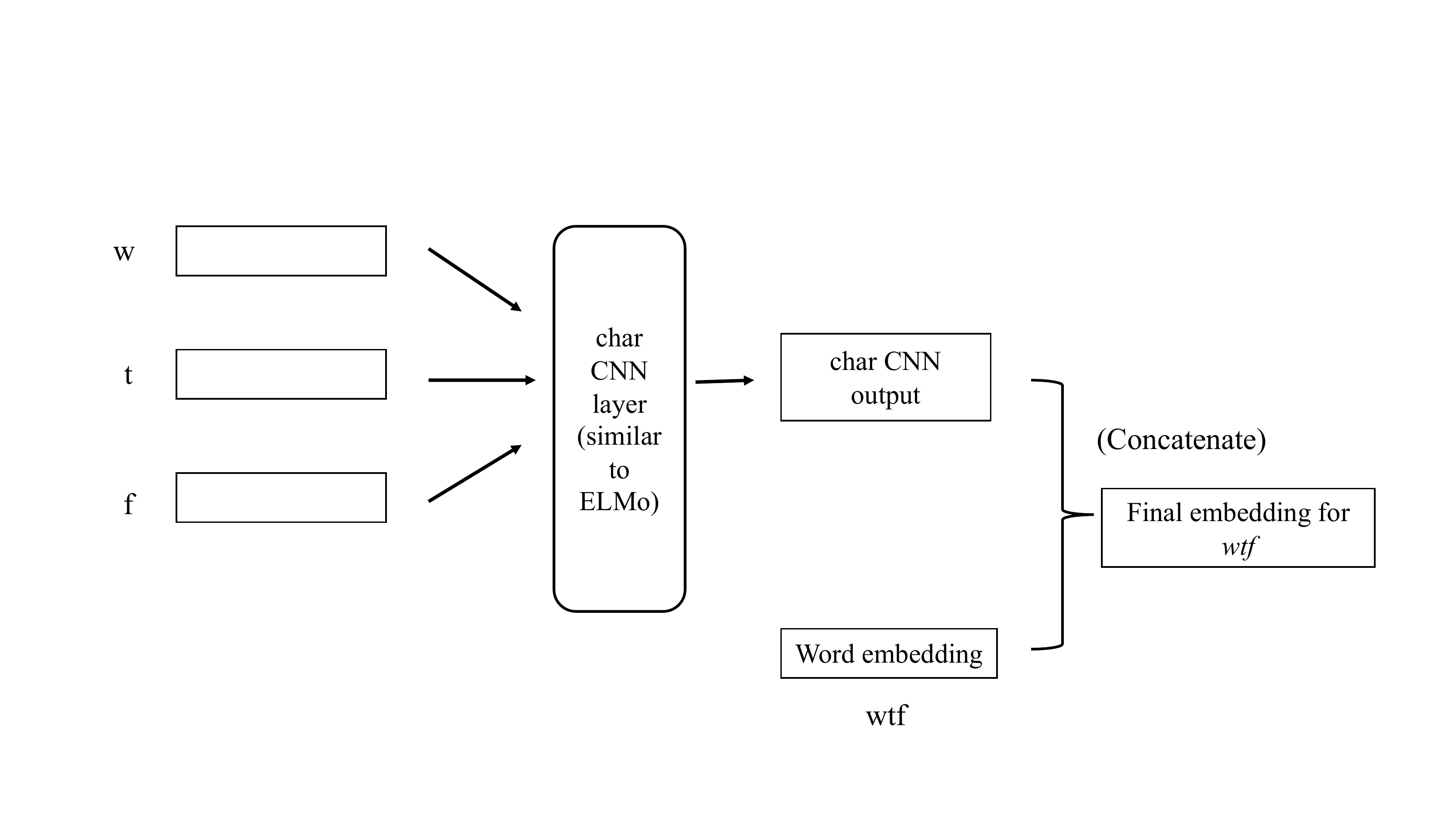}
         \subcaption{}
         \label{fig:char_and_word_emb}
     \end{minipage}
     \hfill{}
     \begin{minipage}[r]{1.0\columnwidth}
         \centering
         \includegraphics[trim=2.5cm 8cm 1.5cm 4cm,clip,width=\textwidth]{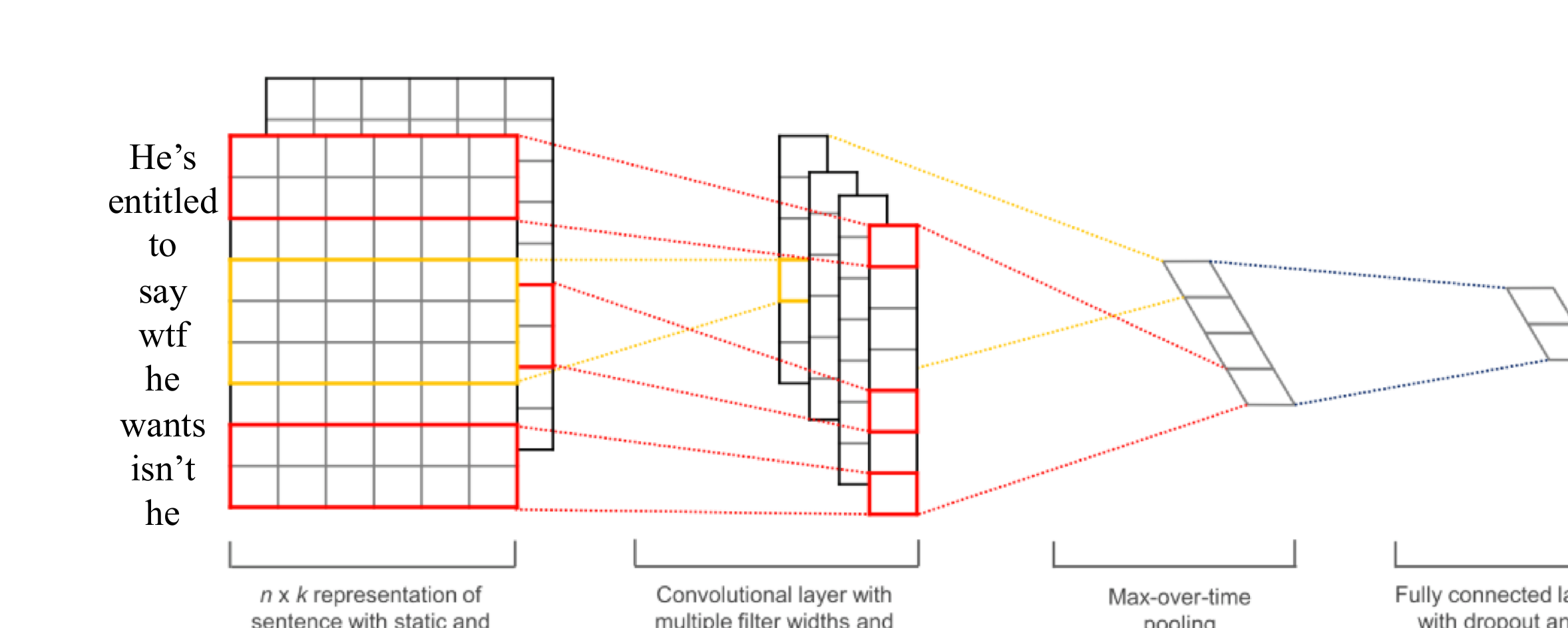}
        \subcaption{}
         \label{fig:kim2014_with_char_emb}
     \end{minipage}
     \caption{Architecture for model described in \ref{ssec:word-char-embeddings}. In Figure \ref{fig:char_and_word_emb}, we present an example of for obtaining a word embedding by concatenating character embeddings with the embedding of the word itself. These final embeddings are then fed into the non-static variant of the Kim2014 \cite{kim2014convolutional} architecture (shown in Figure \ref{fig:kim2014_with_char_emb}). The layers of Kim2014 model alongwith the character CNN layer are updated during training.}
\end{figure*}

Our joint word and character embedding based model is adapted from \citet{kim2014convolutional} and \citet{elmo}.  We refer to \citet{kim2014convolutional} as TextCNN going forward.

 Let $x_i$ be the input word and $c_0^n$ be its character representation, where $n$ is the number of characters in the word.
We transform $c_0^n$ representation by passing through a character embedding layer, which is a n-gram Character-CNN similar to \cite{peters2019tune}. The output of the n-gram CharacterCNN is concatenated with the word's corresponding pretrained embedding to obtain  $w_{full}^{emb}$ as described in \ref{fig:char_and_word_emb} 
Character-level features are concatenated with $w_i^{emb}$ , the word embedding of word $i$, to form $w_{full}^{emb}$, the full set of word-level input features: 
\begin{equation*}
    w_{full}^{emb} = (w_i^{char};w_i^{word})
\end{equation*}
We randomly replace singleton words with special [UNK] (unknown) tokens for obtaining its $w_i^{emb}$ , and also apply dropout \cite{srivastava2014dropout} on $w_{full}^{emb}$.
The input word embeddings $w_{full}^{emb}$, in a sentence with $l$ tokens and convolutional window size $h$, $w_{i:i+h}^{emb}$ is transformed through a convolution filter $w_c$:
\begin{equation*}
 c_i = f(w_c.w_{i:i+h-1}^{emb} + b_c) 
\end{equation*}
where $b_c$ is a bias term and f is a non-linear function (ReLU). This produces a feature map $c$, on-top of which we apply a global max-over time pooling.
\begin{equation*}
    c^^  = max(\textbf{c})
\end{equation*}
This process for one feature is repeated to obtain $m$ filters with different window sizes $h$. The resulting filters are concatenated to form TextCNN document representation. The document representation is passed through Softmax layer to obtain classification predictions. 
We also experiment with original version of TextCNN, which is a pure word based model, without the character embedding variant.


\subsection{End-to-end Character Embedding Model}
\label{ssec:vdccn}


To understand the potential of end to end character based models in dealing noisy text,  we use Very Deep Convolutional Neural Network (VDCNN) architecture proposed by \citet{ConneauEACL17} that operates at character level by stacking multiple convolutional and pooling layers that sequentially extract a hierarchical representation of the text. This representation is fed into a fully connceted 
layer which is trained to maximize the classification accuracy on training data.

\subsection{Byte Pair Encoding + Word + Char embedding models}
\label{ssec:bpe-word-char}
We train a Byte Pair Encoding(BPE) based model introduced by \citet{SennrichHB16a} on the given training corpus. We use this trained BPE model on training data to tokenize/encode our documents in training, validation and test data and use 
each BPE unit as a word to learn embeddings. We perform $30,000$ merge operations 
on each training dataset to learn subword or BPE units. 


\subsection{Pretrained Language Models}
\label{ssec:pretrained-language-models}

Recent liteature have shown that transferring knowledge from large pre-trained language models could benefit various NLP tasks either by adding a task specific architecture or by fine-tuning the language model for the end task \cite{elmo,devlin2018bert,peters2019tune}. In this work, we use $BERT$ 
model and we fine-tune the model for each of our train datasets.




\begin{figure*}[t]
    \centering 
    \includegraphics[trim=1cm 2cm 1cm 2.5cm, clip,width=\textwidth, scale=0.5]{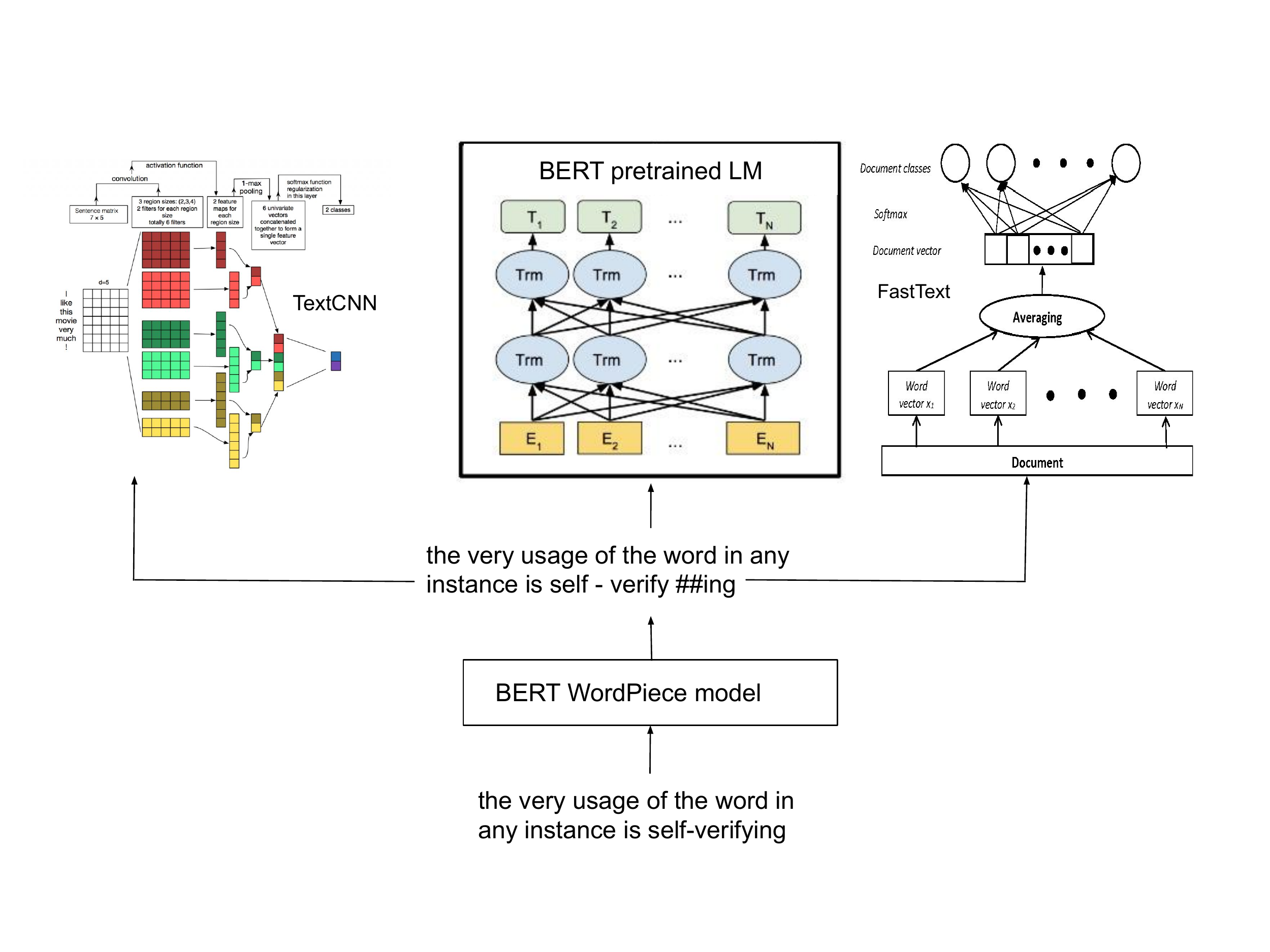}
    \caption{We present the approach discussed in \ref{ssec-bert-bpe-word}. The Input text for a document is tokenized via the BERT Wordpiece tokenized model pretrained on GoogleNews and Wikipedia. This tokenized text is fed as input to the word based models which aids in forming representations from a more informative subword split as an independent unit.}
    \label{fig:bert_pretrained_bpe}
\end{figure*}

\begin{table*}[t]
  \centering
  \tabcolsep 2.5pt
    \begin{tabular}{p{6cm} c c c c c c c c c c c c c} 
     \hline
     \small{\bf{Model}} & \small{\bf{pre}} & \small{\bf{sword}} & \small{\bf{Tok}} && \multicolumn{5}{c}{\small{\bf{\hatespeech}}} 
     && \small{\bf{\wattack}} && \small{\bf{\wtoxic}} \\
     \cline{6-10}
     &&&&&0&1&2&3&4& \\ 
     \cline{2-4} \cline{6-10} \cline{12-12} \cline{14-14}
     fastText$_{ngrams=1}$ & N & N & N && 69.7 & 71.8 & 84.2 & 95.5 & 82.2 && 93.3 && 95.6 \\
     
    
    fastText$_{ngrams=1}$ & Y & N & N && 69.6 & 74.8 & 84.5 & 95.7 & 79.5 && 93.5  &&  95.6 \\
     
    fastText$_{ngrams=1}$ + BERT tokentization   & N & Y & Y && 71.2 & \bf{83.0} & 83.0 & 95.2 & 83.4 && 94.5 && 96.1 \\
    
    fastText$_{ngrams=1}$ + Custom BPE  & N & Y & Y && 66.3 & 72.0 & 74.8 & 73.2 & 72.4 && 81.5  && 84.6 \\
     
    fastText$_{ngrams=2}$ + subword$(2-6)$ & N & Y & N && 64.3 & 71.2 & 75.9 & 92.2 & $85.7^*$ && 93.1 && 95.8 \\
     
    fastText$_{ngrams=2}$ + subword$(2-6)$ \newline + BERT tok  & N & Y & Y && 64.1 & 66.7 & 75.1 & 93.4 & 85.3 && $93.9^*$ && 95.7 \\
     
    fastText$_{ngrams=2}$ + subword$(2-6)$ + + BERT tokentization + preE & Y & Y & N && \bf{71.5} & 76.9 & \bf{87.9} & 93.2 & 75.7 && 93.4  &&  $95.8^*$\\
    \cline{2-14}
     TextCNN \cite{kim2014convolutional} & N & N & N && 69.8 & 76.9 & 85.3 & 95.7 & \bf{85.9} && 92.8  && 95.6\\
     
     TextCNN  + Character n-grams & N & N & N&& 70.6 & 78.1 & 87.1 & 96.3 & \bf{85.9} && 93.2  && 95.9\\
      
     TextCNN + BERT tokenization & N & N & Y && \bf{71.6} & 76.8 & 84.2 & 96.6 & 85.2 && 94.1  && 96.2 \\
    
     VDCNN (9 layers) & N & N & N && 65.3 & 71.6 & 80.7 & 89.3 & 85.9 && 91.6 && 93.9 \\
    
     \cline{2-14}
     BERT (dropout = 0.2) & N & N & N && 72.2 & 80.1 & 85.2 & \bf{97.0} & 78.2  && \bf{95.7} && \bf{96.8} \\

     \hline
    \end{tabular}
    \caption{We report Weighted F1-scores for the different models on the \hatespeech , \wtoxic and \wattack datasets.}
    \label{tab:f1_scores}
\end{table*}

\section{Experiments}
In this section, we present different variants of the models described in Section~\ref{sec:methods}
presented in Table ~\ref{tab:f1_scores}. 

\paragraph{fastText:} \label{ssec-fasttext-subword} We use multiple variants of fastText model. Our fastText$_{ngrams=1}$ uses embeddings learned for each unigram. We treat this as our baseline model without any preprocessing of the text. Our fastText$_{ngrams=2}$ model also uses bigrams along with unigrams as independent tokens to learn embeddings. All pairs of bigrams are chosen wtihout ant frequency threshold. Our fastText$_{ngrams=2}$ + subword $(2-6)$  also uses all subwords within a word boundary within the range of $2-6$. All our models are trained with learning rate of $0.5$ and for $5$ epochs.




\paragraph{TextCNN \cite{kim2014convolutional}:} We run the TextCNN for classification in non-static mode, with learning rate of $0.0001$, dropout of $0.5$ for $50$ epochs. We have used default kernel window sizes $N_f = (3, 4, 5)$ with $m=100$ filters. We initialize the embeddings layer with word2vec pretrained embeddings\footnote{https://code.google.com/archive/p/word2vec/} publicly available from google. We used the non-static variant of TextCNN, with pretrained embedding initialization for word embedding layer. 

\paragraph{TextCNN + char n-grams:} The word embedding layer is constructed for this approach as mentioned in \ref{fig:kim2014_with_char_emb}. The kernel window sizes $h$ for character tokens are $N_f = (1, 2, 3, 4, 5, 6)$ with $m = (32, 32, 32, 64, 64, 64)$ filters respectively.   Increasing the number of filters further to match those of parameters in \citet{elmo} for character tokens led to overfitting on our datasets, and hence we reduced the parameters.  All the layers are allowed to be tuned while training. The character embeddings CNN layer is initialized randomly with Xavier initialization \cite{glorot2010understanding}. We set the character embedding layer output to $300$, upon concatenation the word embedding $w_{full}^{emb}$ length would be $600$. This model is trained in exactly similar settings as the above mentioned word based TextCNN model.


\paragraph{Fully Character Embeddings Model:} We run VDCNN \cite{ConneauEACL17} with $9$ convolution layers with learning rate of $0.0001$ reducing the learning rate by hald every 15 intervals for 100 epochs. We use a batch size of 64 and use stochastic gradient descent (SGD) as optimizization function with $0.9$ momemtum.

\paragraph{BERT:} For our BERT experiments we use the $BERT_{base}$ (uncased) model.
$BERT_{base}$ model consists of 12 Transformer layers with 12 self-attention heads with 768 hidden dimensions and consists of 110 M total parameters. This model is trained in BookCorpus and English Wikipedia corpus. We attach a linear layer on top of $BERT_{base}$ model and the [CLS] token representation is fine-tuned on the training set. We use a binary cross-entropy loss to fine-tune BERT for our datasets. The fine tuned model is evaluated on the test set. We experimented with dropout values set at $(0.1, 0.2)$ between the  transformer encoder layers. We achieved best results at dropout of $0.2$, which we report in our experiments.


\subsection{BERT Wordpiece Tokenizer Model with Word models} \label{ssec-bert-bpe-word}
We use Wordpiece (BPE) model of BERT \cite{devlin2018bert} pretrained on BooksCorpus and English Wikipedia, produced using 30000 merge operations. BERT uses this model as precursor before encoding the text through transformer. We try to examine the benefit of the wordpiece text encoding vs the benefit we obtain from fine-tuning the pretrained LM.  We hypothesize that pretrained BPE model splits a word into most frequent subwords found in the wikipedia corpus, which can help in mining the informative subwords. The informative subwords might prove very beneficial in noisy settings 
where we observe missing spaces and typos. In order to achieve this, we use this pretrained 
BPE model for encoding the document text before inputing to our word based models, TextCNN and fastText word variant. This is demonstrated in Figure~\ref{fig:bert_pretrained_bpe}. We have tried following variants with BERT Wordpiece tokenization as preprocessing step.

\paragraph{BERT Tokenizer with fastText$_{ngrams=2}$ \& TextCNN Word model:} We preprocess the given dataset text using pretrained BPE model, and run a fastText bigram classification model on the preprocessed output. We also evaluate the TextCNN word model with the preprocessed text as input.

\paragraph{BERT Tokenizer with fastText subword:} The preprocessed dataset with BERT trained BPE for training fastText subword model as described in Section \ref{ssec-fasttext-subword}.

\paragraph{Custom BPE model on the dataset:} We also tried to examine if we would get a similar performance boost we obtained from BERT Wordpiece model by encoding text via a custom wordpiece model trained on the text. This helps us differentiate if the gains are from training a wordpiece model on a large text or if the gains are from using subword splitting. We used 30,000 number of merge operations for the custom  BPE model, which is the same as in BERT BPE to aim for a meaningful comparison. We have also tried other values of merge operations from the custom BPE model, but none have yielded substantially better performance.



\begin{table*}[t]
\centering
\begin{tabular}{|p{2cm}|p{13cm}|}
\hline
  \textbf{Technique} & \textbf{PROCESSED DOC}\\
  \hline
  Original &  a complaint about your disruptive behavior here \newline  : https : / / en . wikipedia . org / wiki / wikipedia : administrators \newline  \% 27\_noticeboard / incidents \# disruptive\_users\_vandalizing\_article\_about\_spiro\_koleka \\
  \hline
   Custom BPE  &  complain about your disruptive behavior here \newline  : https : / / en . wikipedia . org / wi@@ ki / wikipedia : administrators \newline \% 27\_noticeboard / incidents \# disrup@@ ti@@ ve\_@@ user@@ s\_@@ \newline vandali@@ z@@ ing\_@@ article\_@@ about\_@@ spi@@ ro@@ \_@@ ko@@ le@@ ka \\
   \hline
   BERT tokentization &  complain about your disrupt @@ive behavior here  \newline  : https : / / en . wikipedia . org / wiki / wikipedia : administrators \newline  \% 27 \_ notice @@board / incidents \# disrupt @@ive \_ users \newline  \_ van @@dal @@izing @@\_ article @@\_ about @@\_ sp @@iro @@\_ ko @@le @@ka \\
   \hline
\end{tabular}
\caption{Sample document split created by BERT BPE tokenizer, Custom BPE tokenizer}
\label{qualitative}
\end{table*}

\section{Results and Analysis}
Table 1 presents the Weighted F1 score based on the support of each of the classes in the test set for our classification task. For a classification problem with $N$ samples in the test set and $C$ classes, Weighted F1 score \footnote{we use sklearn library for computing macro and weighted f1 scores in the paper \url{https://scikit-learn.org/stable/modules/generated/sklearn.metrics.f1_score.html}} is defined as 
\begin{equation}
    \dfrac{1}{N} \sum_{i=1}^{C}n_i*F_i
\end{equation}
where $n_i$ denotes the  number of samples in class $i$. We have reported weighted F1 as the twitter data we obtained had only 17 samples for racism, with stratified CV split having only 4 samples on average. As the results on this label could be very random and prone to lot of variance due to very little number of samples in the train and test set, we choose to use weighted $F1$ over macro $F1$. We also have observed very high variance among performance in different CV splits, hence report the numbers separately on each of them.

\begin{table}[t]
  \tabcolsep 2.5pt
    \begin{tabular}{p{5cm} c c } 
     \hline
    Model & \wattack & \wtoxic \\
    \hline
    WS \cite{mishra2018neural}  & 84.4 & 85.4 \\
    CONTEXT HS + CNG \cite{mishra2018neural}  & 87.4 & 89.3 \\
     \hline
    fastText(ngrams=2) & 85.2 & 86.8 \\
     
    fastText(ngrams=2, BERT BPE)  & 85.9 & 88.6 \\
     
    fastText(ngrams=2, BERT BPE, PreE) & \bf{86.8} & 88.6 \\
    
    Kim2014 & 82.7 & 88.4 \\
     
    Kim2014 (BERT BPE) & 83.4 & \bf{89.3}\\
    \hline
    BERT (dropout = 0.2) & \bf{89.5} & \bf{90.6} \\
     \hline
    \end{tabular}
    \caption{Macro F1 average on the \wtoxic and \wattack datasets.}
    \label{tab:macro_f1_scores}
\end{table}


Table \ref{tab:f1_scores} also mentions if each of the experiment involves using word splitting via BPE, either by pretrained BERT Wordpiece tokenization model, or by training a custom BPE model on our given dataset. We have also highlighted the individual best performance from a modeling architecture with a $*$.



Table 2 presents the Macro $F1$ score on \wattack and \wtoxic datasets. Macro $F1$ score is defined as :
\begin{equation}
    \dfrac{1}{C} \sum_{i=1}^{C}F_i
\end{equation}

\begin{table*}[t]
\centering
\begin{tabular}{|p{2cm}|p{2cm}|p{10cm}|}
\hline
  \textbf{Predicted Label} & \textbf{Technique} &\textbf{Text}\\
  \hline
  $not\_attack$ & Original &  believe that he was the greatest mother-fucker in the world \\
  $attack^*$ & BPE & believe that he was the greatest mother\#\# -\#\# fuck\#\# er in the world \\
  \hline
   $not\_attack$ & Original &  many thanks for your leaving all edits alone in future with such idiotic diatribes \\
  $attack^*$ & BPE &  many thanks for your leaving all edit\#\# s alone in future with such idiot\#\# ic\#\# dia\#\# tri\#\# bes \\
   \hline
\end{tabular}
\caption{Qualitative samples from original text, and BERT Wordpiece model text. Actual label is marked with an asterisk. We can observe that BERT BPE model can effectively mine informative subwords as observed in general domain wikipedia}
\label{qualitative}
\end{table*}

We have picked the best performing models from ~\ref{tab:f1_scores} for macro $F1$ comparison. We have also compared to previous approaches that have achieved best performance on these datasets. 
\citet{mishra2018neural} reported Macro F1 on both validation and test data together. 
From their work it is unclear if the model is tuned on validation, and same data was used along with test to report numbers. Hence, we only use their number as reference.
The main conclusions of these experiments are fourfold:

\paragraph{1. Pretrained BPE models transfer well:}
Pretraining a  Wordpiece model on a large general corpus like wikipedia, and using this for encoding input text by splitting words has shown significant improvements for all the word based models. The fastText word model with bigrams (row 3 in table 1) trained with BERT tokenization achieves the best performance on 1st split of the hatespeech data, and also shows improvement over the native fastText bigrams model on Wiki-ATT dataset. The same observation can be made with TextCNN word model with preprocessing by pretrained BERT Wordpiece tokenization model(row 11 in Table 1). However, we have either noticed a slight degradation or an insignificant improvement by applying BPE encoding with fastText subword based model. This is expected as breaking the informative subwords from BERT into much smaller units might result in lot of noisy updates.

\paragraph{2. Fine tuning pretrained language models:} We observe that fine-tuning large pretrained language models achieve best performance on toxicity dataset. BERT with dropout=0.2 achieves the best performance on most of the datasets and splits. It achieves better or at par performance over any word based model. Only fastText subwords and textCNN/fastText word based model trained on BERT Wordpiece tokenization preprocessing achieve higher performance compared to BERT finetuning.  The gains from BERT Wordpiece tokenization model encoding to fastText word model outperforms performance of BERT model itself. We leave it as future work to further investigate the contribution from BPE Wordpiece tokenization to other classification tasks.

\paragraph{3. End to End Char models arent as effective as subword or word + char models:} Adding character based embedding to aid word embedding based models, and subword models enhance the performance over their pure word based modeling baselines. This proves the hyptohesis of modeling at subword level definitely is beneficial for detecting abusive language. Interestingly, end to end character models arent as effective, which demonstrates the basic fact – knowledge of word leads to a powerful representation, and word boundary information is still informative in noisy settings.

\paragraph{4. State-of-the-art performance on W-TOX and W-ATT with BERT finetuning:}
Table \ref{tab:macro_f1_scores} shows the results for Macro $F1$ score of our models in comparison to previous approaches that have achieved best performance on these datasets.  \citet{mishra2018neural} reported Macro F1 on both validation and test data together. From their work it is unclear if the model is tuned on validation, and same data was used along with test to report numbers. Hence, we only use their number as reference. We have also observed better numbers with their approach. We have achieved state of the art macro $F1$ score on \wattack and \wtoxic datasets with BERT finetuning. We have also added performance of BERT Wordpiece tokenized text with word based models for comparison, with their numbers running really close to those of BERT.

\paragraph{5. Effect of custom BPE model trained on the dataset:}
We have noticed significant performance degaradation as reported in Table \ref{tab:f1_scores}, by tokenizing the text with custom BPE model trained on the W-ATT and W-TOX corpus, in comparison to using the original text or the BERT BPE encoded text. It's interesting to notice the text tokenized by BERT yields very informative subwords, that can help the word based model in comparison to subwords yielded by custom BPE model, even though the vocabulary size of both the models is very similar. Table \ref{qualitative} presents a qualitative example on how the BERT BPE mines informative subwords compared to the custom BPE model. One can note that BERT BPE model clearly splits the text on underscores \& extracts stem of the word in few cases.

\section{Qualitative Analysis}
Table \ref{qualitative} represents couple of examples from W-ATT dataset, where the pure word based model has failed to detect abusive language, but the model trained and tested on BERT Wordpiece tokenized text is able to detect the $attack$. As we can see, Wordpiece model trained on large wikipedia text with 30k operations(BERT) doesnt merge or create relatively uncommon word like $idiotic$ from $idiot$ and $ic$. This helps the model to just learn about $idiot$ clearly from training set, and later use this for clear demarcation.

\section{Conclusion and Future Work}

Existing literature has shown the importance of using finer units such as character or subword units to learn better models and robust representations for identifying abusive language in social media. In this work, we explore various combinations of such word decomposition techniques and present experiments that bring new insights and/or confirm previous findings. Additionally, we study the effectiveness of large pretrained language models trained on standard text in understanding noisy user generated text. We further investigate the effectiveness of subword units (``wordpieces'') learned for unsupervised language modeling can improve the performance of bag-of-words based text classification models such as \fastText. We evaluate our models on Twitter hatespeeech, Wikipedia toxicity and attack datasets.

Our experiments demonstrate that encoding noisy text via BERT wordpiece tokenization model before passing it through word-based models (\fastText and TextCNN) can boost the performance of word-based models and achieve state-of-the-art performance. Based on our experiments, we conclude that subword models perform competitively with character-based models and occasionally outperform them. We observe that adding character embeddings to TextCNN model can slightly boost the performance compared to word-CNN models. 

Our experiments on fine-tuning BERT show improvements on both Wikipedia toxicity and attack datasets. We observe that BERT can effectively transfer pretrained information to classifying tweets and user comments despite the domain shift of pre-training on BookCorpus, Wikipedia Text . Future work in this direction could include pretraining BERT on huge collection of social media text, which might further enhance the performance of identifying abusive language on social media text. Recent work by \citet{wiegand-etal-2019-detection} highlights that most of the datasets that study abusive language are prone to data sampling bias and abusive language identification on realistic scenario is much harder with higher percentage of implicit content. A potential future direction would be to explore how pretrained models on generic text could incorporate or handle implicit abuse.

\section{Acknowledgements}
We would like to thank 
Faisal Ladhak for 
reviewing our work, and all anonymous reviewers for their valuable feedback and comments.

\bibliography{references}
\bibliographystyle{acl_natbib}
\end{document}